\documentclass[10pt,conference]{IEEEtran}
\IEEEoverridecommandlockouts
\usepackage{cite}
\usepackage{amsmath,amssymb,amsfonts}
\usepackage{algorithmic}
\usepackage{graphicx}
\usepackage{textcomp}
\usepackage{xcolor}
\usepackage{caption}
\usepackage{tabularx}
\usepackage{subcaption}
\usepackage{pdflscape}
\usepackage{pdfpages}
\usepackage{multirow}
\usepackage{stfloats}
\def\BibTeX{{\rm B\kern-.05em{\sc i\kern-.025em b}\kern-.08em
    T\kern-.1667em\lower.7ex\hbox{E}\kern-.125emX}}

\usepackage[nogroupskip,nonumberlist,nopostdot]{glossaries}
\usepackage{glossary-mcols}
\newacronym{hpc}{HPC}{High Performance Computing}
\newacronym{fea}{FEA}{Finite Element Analysis}
\newacronym{fdm}{FDM}{Finite Differences Method}
\newacronym{fem}{FEM}{Finite Element Method}
\newacronym{fvm}{FVM}{Finite Volume Method}
\newacronym{pde}{PDE}{Partial Differential Equation}
\newacronym{ai}{AI}{Artificial Intelligence}
\newacronym{ksp}{KSP}{Krylov Subspace Methods}
\newacronym{dof}{DOF}{Degrees of Freedom}
\newacronym{ebe}{EBE}{Element-by-element}
\newacronym{mp}{MP}{Mixed-precision}
\newacronym{gpu}{GPU}{Graphics Processing Unit}
\newacronym{mpi}{MPI}{Message-passing Interface}
\newacronym{cpu}{CPU}{Central Processing Unit}
\newacronym{simd}{SIMD}{Single Instruction Multiple Data}
\newacronym{cg}{CG}{Conjugate Gradient}
\newacronym{pcg}{PCG}{Preconditioned Conjugate Gradient}
\newacronym{minres}{MINRES}{Minimal Residual Method}
\newacronym{gmres}{GMRES}{Generalised Minimal Residual Method}
\newacronym{bicg}{BiCG}{Biconjugate Gradient Method}
\newacronym{acg}{ACG}{Adaptive Conjugate Gradient}
\newacronym{spmv}{SpMV}{Sparse Matrix-Vector Multiplication}
\newacronym{spgemm}{SpGEMM}{Sparse Matrix-Matrix Multiplication}
\newacronym{amg}{AMG}{Algebraic Multi-grid}
\newacronym{gmg}{GMG}{Geometric Multi-grid}
\newacronym{dt}{DT}{Decision Tree}
\newacronym{cnn}{CNN}{Convolutional Neural Network}
\newacronym{ml}{ML}{Machine Learning}
\newacronym{dl}{DL}{Deep Learning}
\newacronym{sor}{SOR}{Successive Over-relaxation}
\newacronym{arm}{ARM}{Acorn RISC Machine}

\newacronym{coo}{COO}{Coordinate}
\newacronym{csr}{CSR}{Compressed Sparse Row}
\newacronym{dia}{DIA}{Diagonal}
\newacronym{ell}{ELL}{ELLPACK}
\newacronym{hyb}{HYB}{Hybrid}
\newacronym{hdc}{HDC}{Hybrid DIA/CSR}
\newacronym{hec}{HEC}{Hybrid ELL/CSR}

\newacronym{vtable}{vtable}{virtual function table}
\newacronym{hpcg-bench}{HPCG}{High Performance Conjugate Gradients}
\newacronym{hpl-bench}{HPL}{High Performance LINPACK}
\newacronym{hbm}{HBM}{High Bandwidth Memory}

\newacronym{cv}{CV}{Cross Validation}
\newacronym{svm}{SVM}{Support Vector Machine}
    
\begin{document}

\title{Optimizing Sparse Linear Algebra Through Automatic Format Selection and  Machine Learning}

\author{\IEEEauthorblockN{Christodoulos Stylianou}
\IEEEauthorblockA{\textit{EPCC, The University of Edinburgh}\\
Edinburgh, UK \\
c.stylianou@ed.ac.uk}
\and
\IEEEauthorblockN{Mich\`{e}le Weiland}
\IEEEauthorblockA{\textit{EPCC, The University of Edinburgh}\\
Edinburgh, UK \\
m.weiland@epcc.ed.ac.uk}
}

\maketitle

\begin{abstract}
Sparse matrices are an integral part of scientific simulations. As hardware evolves new sparse matrix storage formats are proposed aiming to exploit optimizations specific to the new hardware. In the era of heterogeneous computing, users often are required to use multiple formats for their applications to remain optimal across the different available hardware, resulting in larger development times and maintenance overhead. A potential solution to this problem is the use of a lightweight auto-tuner driven by \gls{ml} that would select for the user an optimal format from a pool of available formats that will match the characteristics of the sparsity pattern, target hardware and operation to execute.

In this paper, we introduce \emph{Morpheus-Oracle}, a library that provides a lightweight \gls{ml} auto-tuner capable of accurately predicting the optimal format across multiple backends, targeting the major HPC architectures aiming to eliminate any format selection input by the end-user. From more than 2000 real-life matrices, we achieve an average classification accuracy and balanced accuracy of $92.63\%$ and $80.22\%$ respectively across the available systems. The adoption of the auto-tuner results in average speedup of $1.1\times$ on CPUs and $1.5\times$ to $8\times$ on NVIDIA and AMD GPUs, with maximum speedups reaching up to $7\times$ and $1000\times$ respectively.
\end{abstract}

\section{Introduction}

Sparse matrices, since their inception, have become a crucial component of many scientific simulations in computational science and engineering~\cite{mothra,mini-cfd, Colella_2006}. Over the years, more than 70 representations of sparse matrices (sparse matrix storage formats) have been developed, each addressing the different types of matrices and/or the evolution of hardware towards multi- and many-core processors and accelerators~\cite{spmv_gpu_review}. Literature shows that different formats yield different performance profiles with no single format performing optimally across all different kinds of matrices and types of hardware~\cite{axt,csr5,hdc,monakov_ellpack,clSpMV,sell_c_sigma}.

Exploiting the properties of each format and adjusting the underlying data structure of the sparse matrix (instead of adopting a single storage format) to better fit the operation and target hardware provides optimization opportunities that will potentially improve the performance of the operation. Since the matrix is generally unknown during compile-time, this exploitation can only be done at runtime. Previous efforts~\cite{morpheus, sparse_fortran, smat, ginkgo} provide abstractions and mechanisms that effectively allow for runtime switching to the different formats that are supported.

Even though it is crucial to enable some dynamic switching mechanism to be able to change formats at runtime, it is also important to automate the process of selecting the optimal format. To facilitate the selection of the optimal format for a given matrix, target architecture and operation, auto-tuners have been developed (such as~\cite{smat, smater,clSpMV,cnn_spmv}) with a focus on selecting the optimal format for the \gls{spmv}, the operation that often dominates the runtime of computing the solution to linear systems. These contributions demonstrated the impact of optimizing the performance of iterative solvers through automatic format selection on single node multi- and many-core systems.

In this work, we develop an auto-tuning system that is capable of efficiently tuning the performance of the \gls{spmv} kernel on a wide range of state-of-the-art systems across different vendors using \gls{ml} whilst remaining architecture agnostic. The auto-tuning system offers multiple \gls{ml} algorithms each with different costs and selection accuracy. We introduce \emph{Morpheus-Oracle (Oracle)}~\cite{Morpheus_Oracle}, a library that supports automatic format selection by implementing the different auto-tuners as well as the feature extraction routines. A detailed description of \emph{Oracle} is given in Section~\ref{sec:oracle}.

In summary, our contributions are:
\begin{itemize}
    \item We show the distribution of optimal formats for the \gls{spmv} operation over 6 different matrix storage formats across a variety of systems and target hardware, including multi- and many-core processors and accelerators, for more than 2000 matrices.
    \item For the same set of matrices and systems, we quantify the improvement in performance achieved by choosing the optimal format compared to \gls{csr}, a commonly used general-purpose format. The average runtime speedup is up to $\sim2\times$ on CPU backends and up to $\sim10\times$ on GPU backends.
    \item We provide a reusable model generation system with more than 2000 sparse matrices from real applications that users can exploit to train their own models. In addition we train and tune two different \gls{ml} algorithms and compare their performance in terms of their ability to correctly select the optimal format and quantify the overheads introduced by introducing the auto-tuning mechanism relatively to the runtime cost of \gls{spmv} iterations.
    \item We introduce a simple, flexible and extensible auto-tuning library complementing \emph{Morpheus}, the library for dynamic format switching, that provides tuners which can confidently select the optimal format for a range of architectures.
\end{itemize}
\section{Background and Motivation}
\subsection{Motivation}

New storage formats are proposed every time new architectures emerge aiming to exploit optimizations specific to the new hardware. In the era of heterogeneous computing hardware has become more diverse and as a result applications often require the use of multiple formats across the different types of hardware in order to remain optimal. Even-though new formats have been proposed (\cite{axt,sell_c_sigma,csr5} as an effort to mitigate the performance portability issue there is still no single format that would perform optimally across the different sparsity pattern, hardware architectures and operations. As we show in Section~\ref{sec:optimal_performance}, even when a particular format performs well in general, in some cases can severely under-perform resulting in poor runtime performance. This is predominantly noticeable on GPUs where the wrong format can leave the device under-utilized or result in excessive memory requests due to uncoalesced memory accesses. For applications to achieve optimal performance therefore a better solution is to select the optimal format from a pool of candidate formats at runtime.

Experienced users may have a feeling about the choice of the optimal format for the category or type of matrices they frequently used, however a decision such as this one is not trivial to make as it depends on a number of factors. Furthermore, choosing the optimal format by running the available options first can result in significant overheads. \gls{ml} and \gls{ai} have been successful in various optimization tasks ranging from code optimization to model selection\cite{adaptive_spmv}, including the task of selecting the optimal sparse matrix storage format\cite{smat,cnn_spmv}. Adopting a \gls{ml} model has the potential to offer an accurate and low-overhead solution to the problem of automatic format selection, eliminating any requirement for manual format selection input from the user enabling applications to remain optimal across the different types of hardware and sparsity patterns for any operation of interest.

\subsection{Sparse Matrix Storage Formats}\label{sec:formats}
A plethora of sparse matrix storage formats have been developed over the years, with new formats introduced every time new architectures emerge~\cite{spmv_gpu_review}. Each format has different storage requirements, computational characteristics and comes with different interface for modifying and manipulating the entries of the matrix~\cite{segmented_scan}. Such characteristics make the process of determining, in advance, which format will perform best for a particular matrix given an operation and target hardware non-trivial, with multiple factors contributing in such a decision. In this paper, we consider six different storage formats: ~2 general purpose, ~2 specific purpose and ~2 hybrid formats. 

The most basic and well-known formats are \gls{coo} and \gls{csr}. Both formats are considered \emph{general purpose} formats, and are suitable for a wide range of arbitrary sparsity patterns and target architectures, with \gls{csr} usually adopted as the format of choice. \gls{coo} is constructed from three arrays, where each non-zero element is stored with its pair of row and column indices and no guarantees in the ordering of the elements. Similarly, \gls{csr} also stores explicitly the non-zero values and column indices, but compresses the row indices and stores and array of pointers to mark the boundaries of each row instead, effectively imposing a natural ordering across rows.

\emph{Specific purpose} formats on the other hand aim to exploit the characteristics of a specific class of matrices and are usually designed to perform optimally on a particular architecture, such as GPUs. For example, the \gls{dia} format is designed to represent regular sparsity patterns and is a good fit for vector-like processors. \gls{dia} stores the non-zero elements in a two-dimensional array, where each column holds the coefficients of the diagonal of the matrix. In addition, it holds an integer offset array that keeps track of where each diagonal starts. Another example of interest is the \gls{ell} format, which assumes that there are at most $K$ non-zero entries per row. As a result, it uses one two-dimensional array to represent the non-zero elements of the matrix and another one for the column indices of the values. Consequently, the \gls{dia} format is suitable for representing structures that dominate along the diagonals, such as banded matrices, and \gls{ell} for matrices that are structured or semi-structured (i.e have similar number of non-zeros per row). Note that both formats can suffer from excessive padding if the number of diagonals or the number of non-zeros per row is very large.

\emph{Hybrid} formats usually aim to combine the strengths of two formats in order to eliminate some of their weaknesses, such as the excessive padding mentioned before. The first hybrid format of interest is \gls{hyb}, which is a combination of \gls{ell} and \gls{coo}. \gls{hyb} uses a parameter $K_{H}$ that indicates the number of non-zeros per row to be stored in the \gls{ell} portion. For rows with number of non-zeros larger than $K_H$, the surplus of non-zeros is stored in the \gls{coo} portion instead. The second hybrid format of interest is \gls{hdc}. It uses a parameter $N_D$ that represents the number of non-zeros in a diagonal above which the diagonal is considered to be a \emph{``true'' diagonal}. The true diagonals in the matrix are then stored in \gls{dia} format, whilst the remainder are stored as \gls{csr}.

\subsection{Morpheus}
Morpheus\cite{morpheus} is a C++ library that supports the runtime switching of sparse matrix storage formats through \emph{DynamicMatrix}, a single dynamic ``abstract'' format. \emph{Morpheus} provides a transparent mechanism that can efficiently switch to the different formats supported by the \emph{DynamicMatrix} and it currently supports the six formats mentioned in Section~\ref{sec:formats}. In addition, \emph{Morpheus} offers algorithms such as \gls{spmv} for four different backends in order to support most of the key HPC platforms: 1)~Serial (Sequential), 2)~OpenMP (Multi-threaded), 3)~CUDA (NVIDIA GPUs) and 4)~HIP (AMD GPUs). Furthermore, \emph{Morpheus} provides data management routines and enables data transfers across the different memory spaces of the supported backends and adopts the host-device model to support both homogeneous and heterogeneous platforms (i.e platforms with CPU only or CPU+GPU hardware) maintaining the same application source code.

By abstracting the different formats under a single \emph{dynamic} format, encapsulating the internal implementation details and provide a single interface for algorithms across backends, users are left with a simple and intuitive interface that abstracts away the complexities. In this work, we build on top of \emph{Morpheus}'s runtime switching mechanism to enable automatic format selection and completely eliminate the need for format selection input from the user.

\section{Auto-tuning Pipeline High Level Overview}

\begin{figure}[h]
    \centering
    \includegraphics[width=\columnwidth]{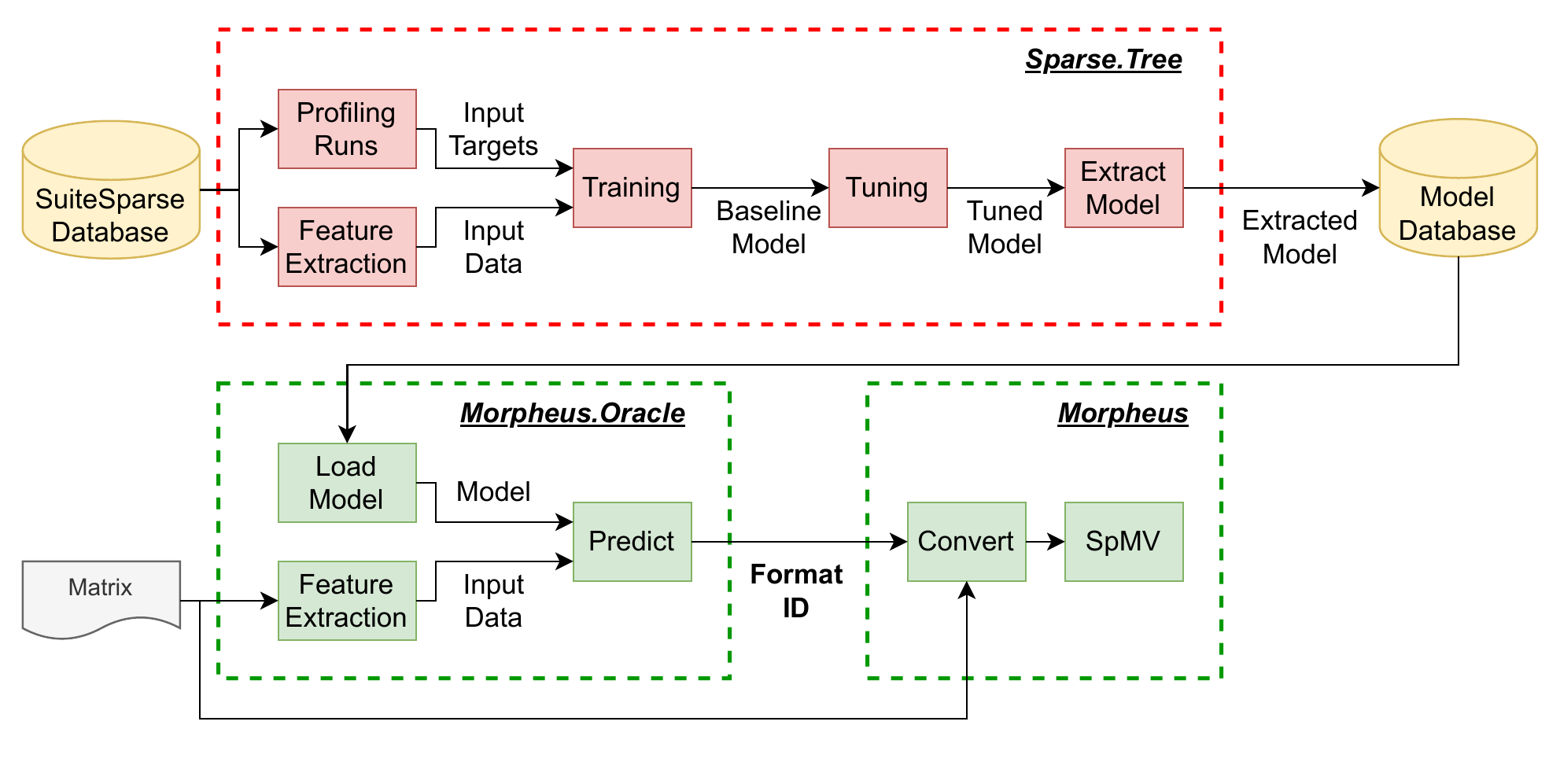}
    \caption{High-level overview of the auto-tuning pipeline. Red and green boxes represent offline and online operations respectively.}
    \label{fig:autotuner_overview}
\end{figure}

The focus of this work is to develop an auto-tuner for selecting the optimal format for the \emph{DynamicMatrix} provided by \emph{Morpheus} to switch to given a matrix, an operation and a target hardware. The most straightforward approach for achieving this task is to utilize a run-first tuner that runs the operation of interest for every format supported, measures the desired metric and selects the best performing format as the optimum. Such an approach will be at expense of the overall runtime performance (even though it will provide the most accurate prediction) as it requires multiple expensive conversions between the different formats, with the expense increasing as more formats are added. A better approach, which reduces the prediction cost, is to use \gls{ml} models to find the optimal format. A high-level overview of our proposed auto-tuning pipeline is shown in Figure~\ref{fig:autotuner_overview}.

The auto-tuning pipeline is divided in the offline (red) and online (green) stage. The offline stage has to be executed once for every new architecture and operation we want to do predictions for, and the results from that stage can then be reused during the online stage.

\subsection{Offline Stage}
The first stage of the pipeline is the actual model generation where we train, tune and extract the \gls{ml} model in a file, for a given architecture and operation, to be used later on by the auto-tuner. We use approximately 2200 real-valued, square matrices of varying sizes, sparsity patterns and different application domains, available from the \emph{SuiteSparse Collection}~\cite{suitesparse}.

For every matrix in the dataset, we first perform profiling runs on the operation and architecture of interest, and measure the runtime in order to determine the optimal format, i.e the format with the shortest runtime for the operation, and export its format ID to be used in the later stages. At the same time, we perform a feature extraction routine, described in detail in Section~\ref{sec:features}, on the matrices in order to generate inputs to be used during the training process. Both the input features (input data) and format ID (input targets) are used to train and tune the \gls{ml} model, a description of which is given in Section~\ref{sec:ml_model}, and once the tuned model is obtained it is exported to a file and stored for later use.

To streamline the training process for users, we wrap this process in a \emph{Python} framework called \emph{Sparse.Tree}\footnote{Available at: https://github.com/morpheus-org/sparse.tree} which uses \emph{scikit-learn}\cite{scikit-learn} under the hood. Users can use \emph{Sparse.Tree} to generate models for new systems or use the pre-trained models from the \emph{Model Database} for the x86 and ARM CPUs or NVIDIA and AMD GPUs used in this work.

\subsection{Online Stage}
In order to be able to automatically select the optimum format to be used by \emph{Morpheus} in an application, we need to be able to make the decision efficiently and online, i.e. while the application is running. In the second stage of the pipeline, we implement \emph{Oracle}, a C++ architecture-independent auto-tuner that uses the \emph{DynamicMatrix} provided by \emph{Morpheus} and loads an \gls{ml} model from a file specified at runtime in order to make the decision. Note that by ``architecture-independent'' we refer to the fact that the tuner is agnostic to the target hardware it is tuning for, as this information is captured by the model loaded at runtime.

For the auto-tuner to be able to use the model, the features of the input matrix need to be extracted in the same way as during the first, training, stage. Then by traversing the model, \emph{Oracle} returns the optimal format ID that \emph{Morpheus} uses to switch to and perform the operation of interest. \emph{Oracle} is described in more detail in Section~\ref{sec:oracle}.

\section{Feature Extraction}\label{sec:features}

Feature extraction in the context of this work refers to the process of transforming the original sparse matrix into a set of numerical ``features'' that can be processed by the model while preserving the information about the sparsity pattern of the original matrix. Relevant features to the problem of interest result in a model that can make informed decisions, however there is a trade-off between the overheads required for computing these features and the accuracy of the decision that is made based on these features. In other words, by providing the model with a large amount of relevant information about the problem will facilitate better learning, but at the cost of having to compute that information. For the purposes of this work, a set of 10 features has been selected (see Section~\ref{sec:selection}) as shown in Table~\ref{tab:features}, capturing information about the basic structure of the sparse matrix but also about the distribution of non-zeros across the rows and diagonals of the matrix. 

\begin{table}[t]
    \centering
    \resizebox{\columnwidth}{!}{%
    \begin{tabular}{c||c|c}
        \textbf{Parameter}    & \textbf{Description} & \textbf{Formula} \\
        \hline
        $M$                             & \# of rows                    & - \\
        \hline
        $N$                             & \# of columns                 & - \\
        \hline
        $NNZ$                           & \# of non-zeros               & - \\
        \hline
        \multirow{2}{*}{$\overline{NNZ}$}    & avg. NNZ                      & \multirow{2}{*}{$\overline{NNZ} = \frac{NNZ}{M}$}\\
                                        & per row                       & \\
        \hline
        \multirow{2}{*}{$\rho$}         & \multirow{2}{*}{density}      & \multirow{2}{*}{$\rho = \frac{NNZ}{M*N}$} \\
                                        &                               & \\
        \hline
        \multirow{2}{*}{$max(NNZ)$}     & max NNZ                       & \multirow{2}{*}{$max(NNZ) = max_{i=1}^{M} NNZ_{i}$} \\
                                        & per row                       & \\
        \hline
        \multirow{2}{*}{$min(NNZ)$}     & min NNZ                       & \multirow{2}{*}{$min(NNZ) = min_{i=1}^{M} NNZ_{i}$} \\
                                        & per row                       & \\
        \hline
        \multirow{2}{*}{$\sigma_{NNZ}$} & std of NNZ                    & \multirow{2}{*}{$\sigma_{NNZ} = \frac{\sum_{i=1}^{M} |NNZ_{i} - \overline{NNZ}|^2}{M}$} \\
                                        & per row                       & \\
        \hline
        $N_D$                           & \# of diagonals               & - \\
        \hline
         \multirow{2}{*}{$N_{TD}$}      & \# of                         & \multirow{2}{*}{-} \\
                                        & true diagonals                & \\
        \hline
    \end{tabular}}
    \caption{Feature parameters used for training the model and, where relevant, the corresponding formula used for computing each one.}
    \label{tab:features}
    \vskip -3mm
\end{table}

\subsection{Feature selection}\label{sec:selection}
The first three features -- number of rows~$(M)$, number of columns~($N$) and number of non-zeros~$(NNZ)$ -- aim to provide a general idea of the size of the matrix, and they are easy to capture as they are provided by the \emph{DynamicMatrix}. According to Monakov et al.\cite{monakov_ellpack}, \gls{coo} is well suited for very sparse matrices with many empty rows and we therefore also add the average number of non-zerors per row~($\overline{NNZ}$) and the density~($\rho$) of the matrix to the set of features.

On the other hand, the performance of specific purpose formats, such as \gls{dia} and \gls{ell}, is heavily affected by the distribution of non-zeros as distributions that are not a good fit for each format will result in excessive padding of zero elements in the matrix hindering the overall performance. \gls{ell} allocates memory based on the maximum number of non-zero elements in a row, therefore matrices with extremely uneven distribution of non-zeros per row are not a good fit for such a format. This information is captured by measuring the maximum and minimum number of non-zeros per row~($max(NNZ), \ min(NNZ)$) and the standard deviation of the non-zeros per row~($\sigma_{NNZ}$). In a similar manner, the \gls{dia} format allocates memory based on the number of diagonals, therefore for matrices that have a large number of diagonals that each only have a few elements will again result in excessive padding. This time therefore the diagonals of the matrix are traversed, keeping count of the number of diagonals~($N_D$) with at least $1$ non-zero and the number of true-diagonals~($N_{TD}$) that have number of non-zeros above a threshold. Since \gls{hyb} and \gls{hdc} are hybrid formats, the existing features remain representative. More details on how the features are actually extracted for the different formats are given in Section~\ref{sec:oracle_feature_extraction}.

\section{Machine Learning Model}\label{sec:ml_model}
Our aim is to train a model that can predict the optimal storage format of a given sparse input matrix. This type of problem falls into the category of multi-class classification problems. During training, for each input matrix in the set, we extract a collection of 10 features and the target attribute that corresponds to the index of the optimal format, obtained from the profiling runs. The objective of the model is to try and determine a mapping between the input features and the optimal format ID, which can be described by Equation~\ref{eq:classification}:
\begin{equation}
\label{eq:classification}
    f(\vec{x_{1}},\vec{x_{2}},...,\vec{x_{n}}) \rightarrow y_{n}(COO,CSR,...,HDC)
\end{equation}
where $\vec{x_{i}}$ represents the feature vector of the $i^{th}$ sparse matrix in the training set and $y_{n}$ represents the target vector with each entry containing the index of one format from the six available. 

To train a model to predict the value of the target of interest,  we are using a decision tree \gls{ml} algorithm that effectively learns simple decision rules inferred from the data features. The reasons for this choice are two-fold: firstly, it is simple to understand and interpret this method; and secondly, it requires little to no data preparation before training the model or using it for prediction. However, decision trees can overfit by creating over-complex trees that do not generalize the data well or become unstable in small variations in the data. To circumvent these issues and generate the optimal model architecture, an exhaustive Grid search is performed to search from the optimal hyperparameter values in a defined hyperparameter space. In addition, to improve the robustness of the model, an ensemble of decision trees is built, called a ``random forest'', that effectively fits a number of decision tree classifiers onto different sub-samples of the dataset. Whilst random forest classifiers can improve the predictive accuracy of the model and control overfitting, this comes at the expense of higher prediction times since multiple trees need to be traversed and the decision from each tree has to be combined into a single final result. 

For this work we are specifically interested in training a model to predict the optimal format to be used during the \gls{spmv} operation for every backend supported in \emph{Morpheus}, however the techniques and algorithms used here are transferable to other sparse operations. Models using both decision tree and random forest methods are trained and tuned in Python, for a number of x86 and ARM CPUs as well as NVIDIA and AMD GPUs, and extracted to a file to be used during the auto-tuning phase.

\section{Morpheus Oracle - An auto-tuner for automatic format selection}\label{sec:oracle}
To facilitate a systematic way of performing format \emph{selection} we developed \emph{Oracle}, a header-only C++ library for automatic format selection. It has been developed to complement the dynamic switching capabilities in \emph{Morpheus} and tune its performance by automating the process of selecting the optimal format to use for a given operation and target architecture. \emph{Oracle} follows a similar design philosophy to \emph{Morpheus}, where containers are separated from the algorithms. Containers here represent the different tuners that are supported by the package and are responsible for encapsulating the specifics of each tuner's implementation exposing the user only to an interface that configures and runs the tuner. 

\subsection{Tuners}
Currently, \emph{Oracle} supports three tuners: 1)~\emph{Run-first}, 2)~\emph{DecisionTreeTuner} and 3)~\emph{RandomForestTuner}. Each tuner is responsible for managing the complexities of selecting the optimal format. For example, the \emph{Run-first} tuner records the iteration time each format takes to perform \emph{N}-iterations for a given operation and applies statistics to determine which format was best. On the other hand, \emph{DecisionTreeTuner} and \emph{RandomForestTuner} require an \gls{ml} model to be loaded from file that is represented using a tree structure which is traversed in order to determine the optimal format. \emph{DecisionTreeTuner} only traverses a single tree whilst \emph{RandomForestTuner} traverses multiple trees in the ensemble and then performs a voting scheme to decide the optimal format. In this case, the majority voting scheme is used that chooses the optimal format to be the one with the most votes. 

The performance of each of the three tuners is a direct trade-off between runtime overhead and prediction accuracy. \emph{Run-first} tuner offers the most accurate prediction at the expense of expensive conversions between each supported format and \emph{DecisionTreeTuner} offers very fast but less accurate predictions. \emph{RandomForestTuner} improves the prediction accuracy of \emph{DecisionTreeTuner} by using multiple trees with the runtime of the prediction process proportional to the number of trees used. The support of multiple types of tuners allows users to choose one based on their requirements.

\subsection{Operations}
Each sparse algorithm offered by \emph{Morpheus} can be tuned to use the optimal format, effectively optimizing the algorithm. An interface for the \gls{spmv} algorithm is defined through the \emph{TuneMultiply} operation, and any additional operations will follow the same principle. By using compile-time introspection we can maintain a single high-level interface for the various supported operations and specialize the implementation for each tuner. Note that the \emph{Oracle} package interacts with \emph{Morpheus} by requesting: 1)~the \emph{DynamicMatrix} to tune for and 2)~the execution space to run the operation in, and it returns the ID of the best format the \emph{DynamicMatrix} should switch to.

The input of the tuning operation requires the \emph{DynamicMatrix} and the tuner, along with the desired  \emph{execution space}, as template parameter. Upon completion of the tuning operation, the tuner can be queried for the optimal format. In the case where the \emph{Run-first} tuner is passed to \emph{TuneMultiply}, the tuning operation will perform \emph{N}-iterations of \gls{spmv} for each format, with the tuner keeping track of the timings. However when \gls{ml} tuners are used, they will evaluate the model and determine the optimal format. In order to be able to do so, \gls{ml} tuners have to perform feature extraction on the fly.

\subsection{Feature Extraction}\label{sec:oracle_feature_extraction}
The biggest challenge when dealing with sparse matrices is that each format has its own representation in memory that can be drastically different between formats. When it comes to feature extraction, the matrix has to be traversed in order to collect the necessary information. Extracting the features defined in Section~\ref{sec:features} on the fly normally would require traversing the rows and diagonals of the matrix multiple times, which for matrices with large number of rows can be expensive. For offline feature extraction during the training process traversing the matrix multiple times might be expensive, but not prohibitive, as the process is generally carried out only once. However, in an online scenario such as during the auto-tuning process this would potentially eliminate any benefits resulting from the optimization process.

To enable format extraction online, \emph{Morpheus} has been extended to provide matrix statistics on a per-format basis and across the different supported backends, eliminating the need for any data transfers across spaces. Multiple statistics can be computed at the same time, reducing the number of times the matrix has to be traversed and reducing the runtime cost of the process. \emph{Oracle} can then perform online feature extraction by inspecting the active format of the \emph{DynamicMatrix} and in turn predict the optimal format by querying the model of the tuner.
\section{Results and Evaluation}
In this section, the performance of the \gls{ml} auto-tuners available in \emph{Oracle} is measured and evaluated. First, profiling runs are carried out to determine the distribution of the optimal format per matrix across the different systems available, and evaluate the performance benefit from switching to the optimal format instead of using a single default format (\gls{csr}). Next, the predictive performance of the \gls{ml} models is quantified comparing the baseline and tuned models. The cost of the prediction process is quantified with respect to the time it takes to carry out a single \gls{spmv} operation and finally the performance of \emph{Morpheus} in conjunction with the auto-tuner by \emph{Oracle} is evaluated.

\subsection{Setup}
All experiments were carried out on the ARCHER2~\cite{archer2_epcc}, Cirrus~\cite{cirrus_epcc} and Isambard~\cite{isambard} supercomputers; their compute node architectures are described in Table~\ref{tab:platforms}.
Experiments run across all four backends supported by \emph{Morpheus} spanning a representative set of all major hardware architectures e.g x86 (Intel and AMD) and ARM CPUs as well as NVIDIA and AMD GPUs.
\begin{table}[h]
    \vskip 3mm
    \begin{center}
    \begin{small}
    \begin{sc}
    \resizebox{\columnwidth}{!}{%
    \begin{tabular}{lllcc}
        \hline
        System & Subsystem & Queue & CPU & GPU \\
        \hline
       \multirow{8}{*}{ISAMBARD} & \multirow{2}{*}{A64FX} & \multirow{2}{*}{A64FX}      & 1x Fujitsu A64FX      & \multirow{2}{*}{-}  \\
                                 &                        &                             & (48 cores)            &                     \\
                                 \cline{2-5}
                                 & \multirow{4}{*}{P3}    & \multirow{2}{*}{Instinct}   &                       & 4x AMD Instinct      \\
                                 &                        &                             &  1x AMD EPYC 7543P    &        MI100         \\
                                 &                        & \multirow{2}{*}{Ampere}     &  (32 cores)           & 4x Nvidia Ampere     \\
                                 &                        &                             &                       & A100 40GB            \\
                                 \cline{2-5}
                                 & \multirow{2}{*}{XCI}   & \multirow{2}{*}{ARM}        & 1x Marvell ThunderX2  & \multirow{2}{*}{-}   \\
                                 &                        &                             & ARM (32 cores)        &                      \\
        \hline
        \multirow{4}{*}{CIRRUS}  &                        & \multirow{2}{*}{Standard}   & 2x Intel Xeon         & \multirow{2}{*}{-}   \\
                                 &                        &                             &  E5-2695 (18 cores)   &                      \\
                                 &                        & \multirow{2}{*}{GPU}        & 2x Intel Xeon         & 4x Nvidia Volta      \\
                                 &                        &                             &  Gold 6248 (18 cores) & V100 16GB            \\
        \hline
        \multirow{2}{*}{Archer2} &                        & \multirow{2}{*}{Standard}   & 2x AMD EPYC 7742      & \multirow{2}{*}{-}   \\
                                 &                        &                             &  (64 cores)           &                      \\
        \hline
    \end{tabular}}
    \end{sc}
    \end{small}
    \setlength{\belowcaptionskip}{-8pt}   
    \caption{Node configurations for the systems used in the experiments.}
    \label{tab:platforms}
    \end{center}
    \vskip -3mm
\end{table}

The dataset used in the experiments consists of approximately 2200 real, square matrices available from the SuiteSparse library, in an 80\%-20\% split between training and test set.

\subsection{Format Distribution}\label{sec:format_distribution}
\begin{figure}[h]
    \centering
    \includegraphics[width=\columnwidth]{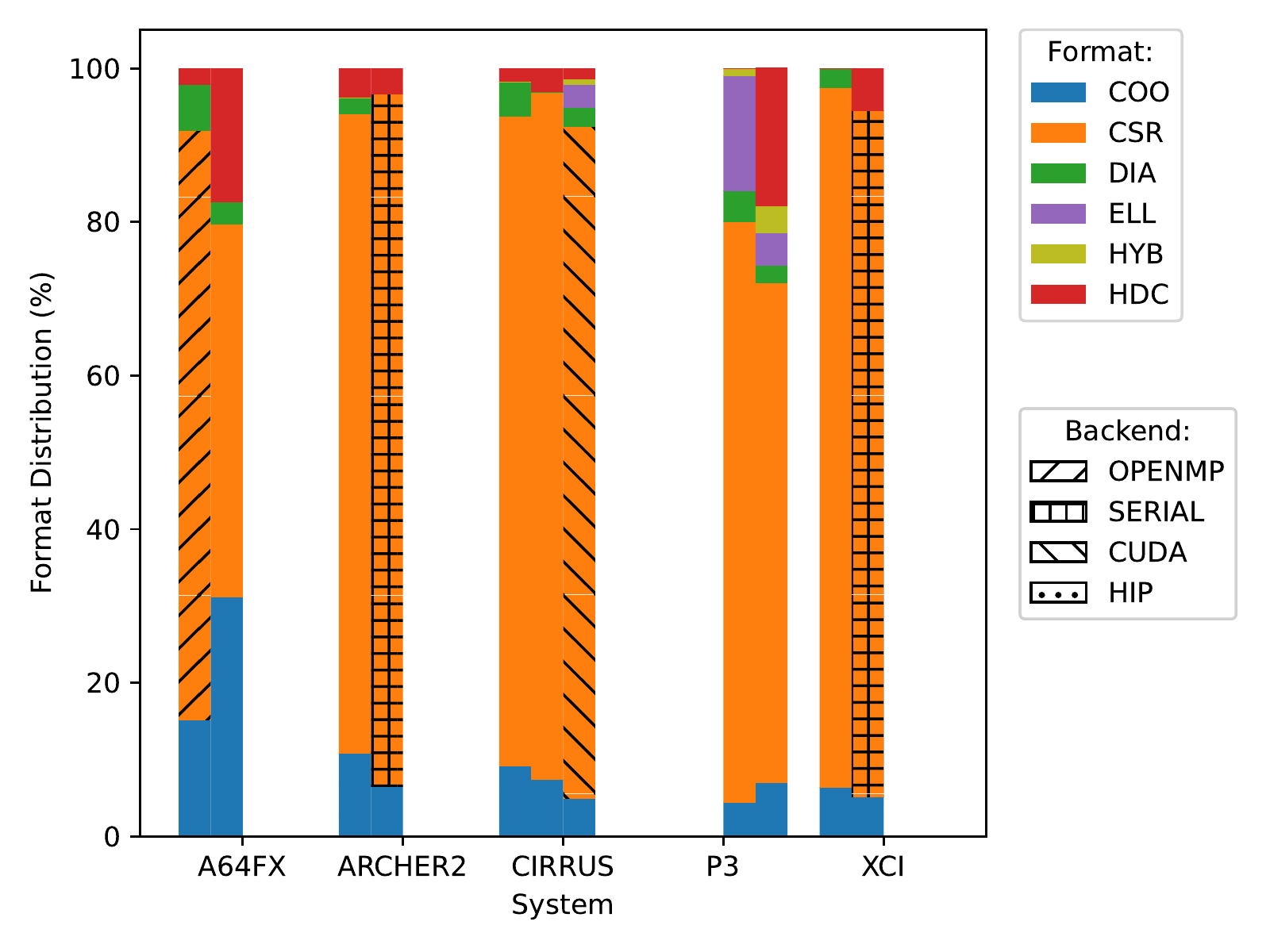}
    \caption{Optimal Format distribution for 1000 repetitions of \gls{spmv} using the SuiteSparse dataset. The optimal format for each matrix is selected to be the one with the smallest runtime.}
    \label{fig:matrix_distribution}
\end{figure}

Since no single format performs best across different sparsity patterns, target hardware and operations, in order to get an understanding of which format performs optimally across the different matrices in the dataset, profiling runs of the \gls{spmv} operation are carried out on the available platforms. For every matrix in the dataset, supported format and available platform the runtime of 1000 \gls{spmv} repetitions is recorded and the format with the minimum runtime is set to be the optimal format for the particular matrix and platform. Figure~\ref{fig:matrix_distribution} shows the distribution of the optimal formats per system and format for all matrices in the SuiteSparse dataset.

From Figure~\ref{fig:matrix_distribution} it is clearly shown that for the biggest portion of matrices in the set the optimal format across systems and backends is \gls{csr}, validating its role as the most commonly used storage format. However, we do observe that even on the same hardware the distribution can change quite drastically. For the Serial backend on the A64FX system more matrices perform better by using \gls{hdc} or \gls{coo} formats, whereas for the OpenMP backend these become \gls{csr}. However, on Cirrus and Archer2 systems the opposite effect is observed where matrices that perform best with \gls{csr} on the Serial backend are actually performing best with \gls{coo} or \gls{dia} on the OpenMP backend. In addition, the distribution on GPU systems compared to the CPU systems is much more diverse with optimal formats chosen from almost every available format class.

The main takeaway here is that the format distribution for every system/backend in the experiment is unbalanced, with \gls{csr} being the clear majority. Therefore, the classification problem in Section~\ref{sec:ml_model} falls in the category of the imbalanced classification problems or rare event prediction. A large portion of real-life matrices perform well using \gls{csr}, however an auto-tuner that can predict rare events is useful if in the case where selecting a different format benefits performance noticeably.

\subsection{Optimal Format Performance}\label{sec:optimal_performance}
\begin{figure}[h]
    \centering
    \includegraphics[width=\columnwidth]{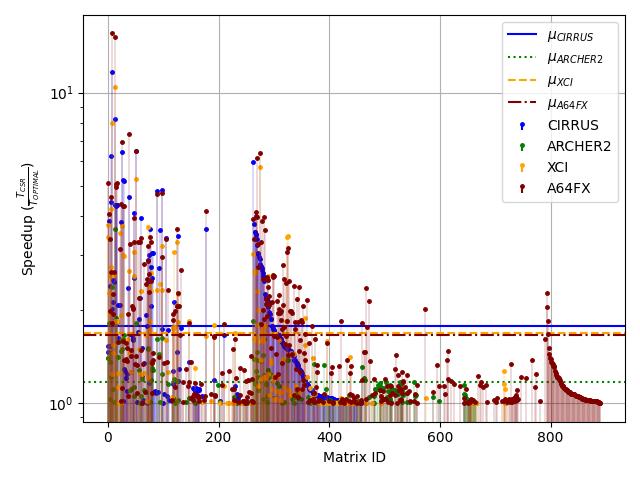}
    \caption{Runtime speedup of \gls{spmv} using the optimal format against \gls{csr} on the OpenMP backend of the available systems for the SuiteSparse dataset. Matrices with optimal format set to \gls{csr} are omitted for clarity.}
    \label{fig:profiling-performance-omp}
\end{figure}

As part of the experiment shown in Section~\ref{sec:format_distribution}, the timings of \gls{spmv} operation were recorded in an effort to quantify the real benefit in those cases where the optimal format is not \gls{csr}. In Figure~\ref{fig:profiling-performance-omp} the runtime of \gls{spmv} using \gls{csr} is measured against the equivalent runtime of the optimal format for each matrix in the dataset on the OpenMP backend and across the available systems. Note that matrices with optimal format set to \gls{csr} are omitted for clarity. Whilst a lot of the matrices result in a speedup of less than $1.5\times$, there is a noticeable number of matrices that exhibit speedups between $1.5\times$ and $10.5\times$, with an average speedup of approximately $1.8\times$ for Cirrus, XCI and A64FX, and of $1.3\times$ on Archer2. Similar results are obtained for the Serial backend on the same systems; these have been omitted in the interest of the available space. 

\begin{figure}[t]
    \centering
    \includegraphics[width=\columnwidth]{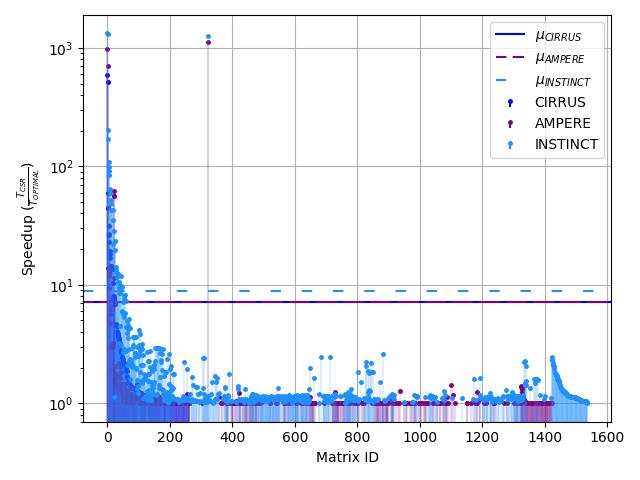}
    \caption{Runtime speedup of \gls{spmv} using the optimal format against \gls{csr} on CUDA and HIP backends. An NVIDIA V100 and A100 GPU is used on Cirrus and Ampere (Isambard) systems and an AMD MI100 on Instinct (Isambard). Matrices with optimal format set to \gls{csr} are omitted for clarity. The average speedup is $8\times$ and $10\times$ on CUDA (both for V100 and A100) and HIP backends respectively.}
    \label{fig:profiling-performance-gpu}
    \vskip -3mm
\end{figure}

For the CUDA and HIP backends, the runtime speedups are more noticeable compared to the ones measured on the CPU backends. As shown in Figure~\ref{fig:profiling-performance-gpu}, the average speedup for the CUDA and HIP backends is $8\times$ and $10\times$ respectively, with maximum speedups reaching up to $1000\times$. After closer inspection of the memory read/write requests, and the occupancy achieved during the launch of the \gls{spmv} kernel, for one of the matrices (\emph{mawi\_201512020030}), the \gls{csr} version issues $5\times$ more requests and the occupancy is $10\times$ smaller compared to the version using the optimal format. The sparsity pattern of the matrix results in uncoalesced accesses and leaves the GPU under-utilized when the \gls{csr} format is used. 

These results justify the development and use of an auto-tuner such as the one proposed in this paper since the choice of format can have noticeable benefits in the performance of the operation. However, the auto-tuner must be lightweight to avoid performance degradation in the case where the optimal format is \gls{csr}.

\begin{table*}[b]
    \centering
    \resizebox{\textwidth}{!}{%
    \begin{tabular}{l|l||c|c|c|c|c|c|c|c|c|c|c|c|c|c|c|c|c|c|}
        \multirow{3}{*}{\textbf{System}} & \multirow{3}{*}{\textbf{Backend}} &  \multicolumn{2}{c|}{\textbf{Estimators}}   & \multicolumn{2}{c|}{\textbf{Bootstrap}}  & \multicolumn{2}{c|}{\textbf{Max}} & \multicolumn{2}{c|}{\textbf{Min}}  & \multicolumn{2}{c|}{\textbf{Min}}  & \multicolumn{2}{c|}{\textbf{Max}} & \multicolumn{2}{c|}{} & \multicolumn{2}{c|}{\textbf{Accuracy}} & \multicolumn{2}{c|}{\textbf{Balanced}} \\
                                &                          & \multicolumn{2}{c|}{}  &  \multicolumn{2}{c|}{} & \multicolumn{2}{c|}{\textbf{Depth}}    & \multicolumn{2}{c|}{\textbf{Samples}} & \multicolumn{2}{c|}{\textbf{Samples}} & \multicolumn{2}{c|}{\textbf{Features}} & \multicolumn{2}{c|}{\textbf{Criterion}} & \multicolumn{2}{c|}{\textbf{(\%)}} & \multicolumn{2}{c|}{\textbf{Accuracy}} \\
                                &                          & \multicolumn{2}{c|}{} &  \multicolumn{2}{c|}{} &   \multicolumn{2}{c|}{}   & \multicolumn{2}{c|}{\textbf{Leaf}} & \multicolumn{2}{c|}{\textbf{Split}} &  \multicolumn{2}{c|}{} & \multicolumn{2}{c|}{}  & \multicolumn{2}{c|}{}  & \multicolumn{2}{c|}{\textbf{(\%)}} \\
        \hline
        \multirow{2}{*}{Archer2} & Serial & \multirow{11}{*}{100} & 40  & \multirow{11}{*}{T} & F & 21 & 21 & \multirow{11}{*}{1} & 3   & \multirow{11}{*}{2} & 2   & \multirow{11}{*}{10} & 4  & \multirow{11}{*}{gini} & entropy  & 95.13   & \textbf{95.86}  & 83.65  & \textbf{88.49} \\
                         & OpenMP &                       & 40  &                     & T &         20            & 14 &                     & 1   &                     & 10  &                     & 9  &                        & entropy  & 91.94   & \textbf{92.18}  & 83.39  & \textbf{85.12} \\
        \cline{1-2}\cline{4-4}\cline{6-8}\cline{10-10}\cline{12-12}\cline{14-14}\cline{16-20}
\multirow{3}{*}{Cirrus}  & Serial &                       & 50  &                     & T &                 18    & 18 &                     & 2   &                     & 2   &                     & 6  &                        & entropy  & 93.60   & \textbf{94.08}  & 79.13  & \textbf{81.71} \\
                         & OpenMP &                       & 30  &                     & T &            19         & 15 &                     & 1   &                     & 10  &                     & 8  &                        & gini     & \textbf{92.65}   & 91.71  & 68.36  & \textbf{78.39} \\
                         & Cuda   &                       & 50  &                     & F &             17        & 16 &                     & 1   &                     & 10  &                     & 4  &                        & entropy  & 92.89   & \textbf{93.60}  & 71.72  & \textbf{74.32} \\
        \cline{1-2}\cline{4-4}\cline{6-8}\cline{10-10}\cline{12-12}\cline{14-14}\cline{16-20}
\multirow{2}{*}{A64FX}   & Serial &                       & 90  &                     & T &               20      & 18 &                     & 1   &                     & 2   &                     & 5  &                        & gini     & 87.93   & \textbf{87.93}  & 84.56  & \textbf{86.22} \\
                         & OpenMP &                       & 30  &                     & T &         19            & 13 &                     & 1   &                     & 5   &                     & 6  &                        & gini     & 91.26   & \textbf{91.75}  & 89.29  & \textbf{91.62} \\
        \cline{1-2}\cline{4-4}\cline{6-8}\cline{10-10}\cline{12-12}\cline{14-14}\cline{16-20}
\multirow{2}{*}{P3}      & Cuda   &                       & 40  &                     & F &              22       & 14 &                     & 2   &                     & 10  &                     & 4  &                        & entropy  & 86.46   & \textbf{87.17}  & \textbf{84.95}  & 83.82 \\
                         & HIP    &                       & 40  &                     & T &          19           & 11 &                     & 1   &                     & 2   &                     & 6  &                        & entropy  & \textbf{93.38}   & 92.67  & \textbf{90.82}  & 87.92 \\
        \cline{1-2}\cline{4-4}\cline{6-8}\cline{10-10}\cline{12-12}\cline{14-14}\cline{16-20}
\multirow{2}{*}{XCI}     & Serial &                       & 60  &                     & F &              24       & 12 &                     & 2   &                     & 10  &                     & 6  &                        & entropy  & 96.21   & \textbf{96.92}  & 91.34  & \textbf{95.72} \\
                         & OpenMP &                       & 20  &                     & T &         16            & 10 &                     & 1   &                     & 5   &                     & 10 &                        & entropy  & 94.54   & \textbf{95.01}  & 55.25  & \textbf{75.31} \\
        \hline
        \hline
        \multicolumn{15}{c|}{} & \textbf{Mean} & 92.36 & 92.63 & 80.22 & 84.42 \\
        \multicolumn{15}{c|}{} & \textbf{Std ($\pm$)}  & 2.93  & 3.02  & 11.04 & 6.64 \\
    \end{tabular}}
    \caption{Hyperparameters used during the tuning process of the Random Forest classifier. The accuracy and balanced accuracy achieved for the test set are also reported. We report both the baseline (left sub-columns) and tuned (right sub-columns) classifier parameters and score. The tuning process uses a 5-fold \gls{cv} method.}
    \label{tab:RandomForestTuning}
\end{table*}

\subsection{Hyperparameter Tuning}\label{sec:hyperparameters}

Hyperparameter tuning for \gls{ml} algorithms is essential for the overall performance of the \gls{ml} model. This process relies more on experimental results rather than theory and the best method to determine the optimal settings is by trying different hyperparameter combinations and evaluate the performance of the model. To account for overfitting and ensure the model generalizes well on unseen data we perform a 5-fold \gls{cv} on the training set and iteratively fit the model $5$ times each time training on $4$ folds and validating on the $5th$.

For hyperparameter tuning, a grid search is performed to search for the optimal hyperparameter values in a defined hyperparameter space, each time performing the entire 5-fold \gls{cv} process. We compare all the generated models and the best one is selected to train on, using the full training set, obtaining the tuned model. In our case, we perform the tuning process for both the \emph{decision tree} and \emph{random forest} classifiers.

Table~\ref{tab:RandomForestTuning} shows the hyperparameters used to generate the baseline (left sub-columns) and tuned (right sub-columns) \emph{random forest} for each available backend and system, along with the achieved accuracy and balanced accuracy on the test set. Similar qualitative results were achieved for the \emph{decision tree} classifier, hence are omitted in the interest of the available space. We tune for the main parameters that affect the generalization ability of a \emph{random forest} classifier (the number of estimators, max depth of the trees and the minimum number of samples on the leaf nodes) and we also test different secondary parameters such as bootstrap (sampling data points with or without replacement), minimum samples required before a split, maximum features considered before splitting a node and the criterion function used to measure the quality of the split. 

Even though the tuning process results in a model that reports a similar average accuracy score ($92.63\%$) to the baseline ($92.36\%$), the tuned model is using significantly fewer and shallower trees (estimators) resulting in much faster prediction times. However, since the dataset is unbalanced, a more indicative metric to report is the balanced accuracy calculated as the average of the proportion of correctly classified samples of each class individually. In this case, the tuned model ($84.42\%\pm6.64\%$) performs noticeably better compared to the baseline ($80.22\%\pm11.04\%$). It is worth pointing out that for some system and backend pairs the change in balanced accuracy is quite drastic (e.g $10\%$ and $20\%$ increase for the XCI and Cirrus systems using the OpenMP backend respectively). 

For reference, the accuracy and balanced accuracy achieved on the tuned \emph{decision tree} is $90.85\%\pm7.87\%$ and $78.12\%\pm4.91\%$ respectively. This result justifies the development of both \emph{DecisionTreeTuner} and \emph{RandomForestTuner} as this allows for a faster predictions when the \emph{DecisionTreeTuner} is used without significant sacrifice in prediction accuracy.

\subsection{Auto-tuner Performance}\label{sec:tuner_performance}
The tuned \emph{random forest} classifier with parameters set as in Table~\ref{tab:RandomForestTuning} is deployed in C++ using the \emph{RandomForestTuner} in \emph{Oracle} on a synthetic benchmark. The benchmark performs 1000 \gls{spmv} operations using a sparse matrix switched to the optimum format selected by the tuner at runtime. The dataset used in the benchmark consists of all the matrices in the test set. For each system and backend pair, the auto-tuner runtime performance is measured in the form of equivalent \gls{spmv} operation using the \gls{csr} format, given by $T_{tuning} = \frac{T_{CSR}}{T_{FE} + T_{PRED}}$, where $T_{CSR}$, $T_{FE}$ and $T_{PRED}$ are the runtime of a single \gls{csr} \gls{spmv}, feature extraction and prediction operations respectively. 

\begin{table}[h]
    \centering
    \resizebox{\columnwidth}{!}{%
    \begin{tabular}{l|l||c|c|c|c|c|c|c|}
     \textbf{System}             & \textbf{Backend} & \textbf{Mean} & \textbf{Std} & \textbf{Min} & \textbf{Q1} & \textbf{Q2} & \textbf{Q3} & \textbf{Max} \\
\hline
\multirow{2}{*}{Archer2}    & Serial           & 10            & 19           & 2           &  4          &  7          & 10          & 303         \\
                            & OpenMP           & 25            & 20           & 2           & 14          & 21          & 31          & 179         \\
\hline
\multirow{3}{*}{Cirrus}     & Serial           & 10            & 30           & 2           &  3          &  4          &  7          & 359         \\
                            & OpenMP           & 64            & 72           & 2           & 21          & 33          & 83          & 643         \\
                            & CUDA             &  7            &  3           & 1           &  6          &  6          &  8          &  29         \\
\hline
\multirow{2}{*}{A64FX}      & Serial           &  6            &  9           & 1           &  3          &  4          &  5          & 120         \\
                            & OpenMP           & 45            & 40           & 1           & 23          & 28          & 59          & 246         \\
\hline
\multirow{2}{*}{P3}         & CUDA             &  2            &  3           & 1           &  2          &  2          &  2          &  42         \\
                            & HIP              & 15            &  9           & 1           &  7          & 18          & 23          &  30         \\
\hline
\multirow{2}{*}{XCI}        & Serial           & 12            & 28           & 2           &  6          &  7          &  9          & 335         \\
                            & OpenMP           & 17            & 29           & 2           &  3          &  6          & 18          & 203         \\
\hline
\hline
    \end{tabular}}
    \caption{The runtime cost, expressed in terms of \gls{spmv} operations using \gls{csr}, of using the auto-tuner. Measured as $T_{tuning} = \frac{T_{CSR}}{T_{FE} + T_{PRED}}$, where $T_{CSR}$, $T_{FE}$ and $T_{PRED}$ are the runtime of a single \gls{csr} \gls{spmv}, feature extraction and prediction operation respectively. Columns show the statistics of the runtime cost and Q1 to Q3 represent the quartiles.}
    \label{tab:IterationCount}
\end{table}

Table~\ref{tab:IterationCount} shows statistics on the runtime cost of the tuner in the form of the number of \gls{spmv} operations using \gls{csr} for every system and backend pair. We do observe that the OpenMP backend requires on average more time to run the auto-tuner irrespective of the system, even-though it might use fewer estimators compared to the equivalent Serial backend. This indicates that the \emph{random forest} using the OpenMP backend consists of fewer but more complex estimators compared to Serial. \emph{RandomForestTuner} on GPUs on the other hand spends less time running the auto-tuner as the average runtime cost is only few repetitions compared to CPU backends. At least $75\%$ of the matrices in the test set require fewer than 100 repetitions for the tuning process. In real-life applications, for example solving a time-dependent \gls{pde}, would require many thousands of \gls{spmv} operations meaning that the auto-tuner proposed would not incur noticeable overheads. Even in the maximum cases observed, the cost of running the tuner remains within accepted limits. 

\subsection{Tuned SpMV Performance}

Using the same setup as in Section~\ref{sec:tuner_performance} we also quantify the runtime speedups in \gls{spmv} obtained by adopting the proposed auto-tuner compared to \gls{spmv} using \gls{csr}. The speedup is measured as shown in Equation~\ref{eq:tuner_speedup}:
\begin{equation}\label{eq:tuner_speedup}
    Speedup = \frac{T_{CSR}}{T_{TUNE} + T_{OPT}} = \frac{T_{CSR}}{T_{FE} + T_{PRED} + T_{OPT}}
\end{equation}
where $T_{CSR}$, $T_{OPT}$, $T_{FE}$ and $T_{PRED}$ are the runtime of 1000 \gls{spmv} operations using \gls{csr} and the predicted format, feature extraction and prediction operations respectively.

\begin{figure*}[t]
    \centering
    \captionsetup{justification=centering}
    \begin{subfigure}[h]{0.49\textwidth}
            \captionsetup{justification=centering}
            \includegraphics[width=\columnwidth]{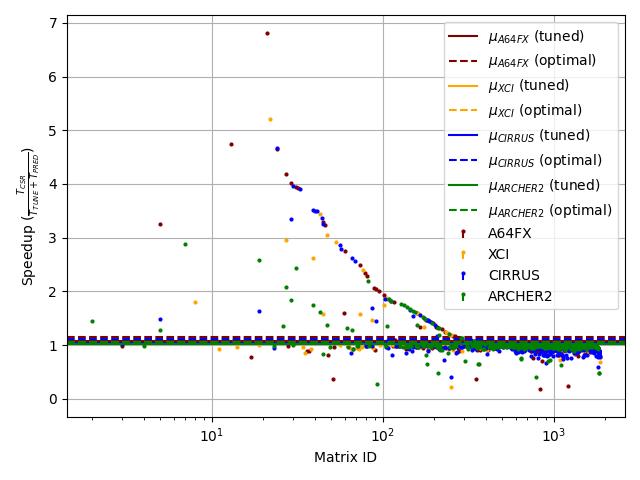}
    \caption{OpenMP}
    \label{fig:rf-performance-omp}
    \end{subfigure}
    ~
    \begin{subfigure}[h]{0.49\textwidth}
            \captionsetup{justification=centering}
            \includegraphics[width=\columnwidth]{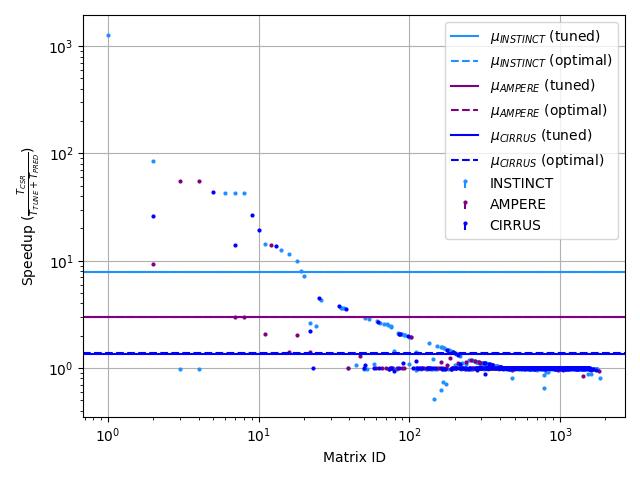}
    \caption{CUDA/HIP}
    \label{fig:rf-performance-gpu}
    \end{subfigure}
    \caption{Obtained runtime speedup from using the auto-tuner and predicted format against using \gls{csr} in performing 1000 \gls{spmv} operations on the available systems for every matrix in the test set. The average speedup ($\mu$) from using the predicted format (includes also the time the tuner requires to make a prediction) matches the average speedup from using directly the optimal format, indicating that minimal overheads are introduced on average from using the tuner.}
    \label{fig:rf-performance}
\end{figure*}

Figure~\ref{fig:rf-performance} shows the achieved speedup on OpenMP, CUDA and HIP backends on the available systems for all the matrices in the test set. Note that the Serial backend is excluded due to space constraints as a similar trend to the OpenMP backend is observed. On CPUs, the speedup from introducing the auto-tuner and selecting the predicted format results in similar performance as if we were to use the \gls{csr} format. The result is consistent across all available systems with average speedup close to $1.1\times$. For the majority of matrices the overheads from the auto-tuner do not reduce overall performance as most of the samples are concentrated around $1$. The few for which performance falls significantly below $1$, we do observe the impact from wrongly classifying the optimal format. However, in many cases the auto-tuning process results in noticeable speedups, with maximum achieved speedup of $7\times$ on the A64FX system.

On the GPU backends (Figure~\ref{fig:rf-performance-gpu}), the auto-tuning approach is much more beneficial compared to the OpenMP backend as higher average speedups are achieved. On average, for the NVIDIA A100 (Ampere) and V100 (Cirrus) GPUs a $1.5\times$ and $3\times$ average speedup is achieved respectively, a result that follows the bandwidth ratio between the two architectures. The AMD M150 (Instinct) GPU however is the one architecture that clearly benefits from the introduction of the auto-tuner as it reports an average speedup of $8\times$. Note that compared to the CPU backends, on GPUs the performance of a mis-classification is less severe as fewer samples fall significantly below $1$. In addition, for a number of matrices the achieved speedup improves performance by orders of magnitude highlighting the importance of adopting an auto-tuning approach for format selection. In all three backends shown in Figure~\ref{fig:rf-performance} the average speedup achieved from using the auto-tuner matches the average optimal speedup for when the optimum format was selected (without performing any auto-tuning) suggesting that the overheads introduced by the auto-tuner become negligible as the number of \gls{spmv} repetitions increases.
\section{Related Work}
Over the years \gls{ml} has been proven a valuable approach for performing various optimization tasks such as code optimization, task scheduling and model selection\cite{adaptive_spmv}. Applying auto-tuning techniques for optimizing sparse linear algebra remains an active area of research with developments spanning topics from format specific parameter tuning (\cite{adaptive_spmv,sell_c_sigma,axt}) to automatic format selection across current and emerging architectures.

Benatia et al.\cite{svm_selection} proposed an \gls{svm} classifier for selecting the optimal format from four available formats for the \gls{spmv} on GPUs, reporting classification accuracy up to $88\%$. Similarly, Sedeghati et al.\cite{dt_selection_gpu} used a Decision Tree based classifier to choose from five available formats resulting in $81\%$ accuracy on GPUs. On the other hand, Li et al.\cite{smat} used an input adaptive \gls{spmv} auto-tuner based on ruleset classification that maintains a confidence value for each test sample. If the prediction of the classifier is below the defined threshold, the auto-tuner selects the optimal format and \gls{spmv} kernel otherwise switches to a run-first approach to make the decision, reporting accuracy of up to $85\%$.

To alleviate the problem of manually defining the features to be extracted, Zhao et al.\cite{cnn_spmv} uses a \gls{cnn} model for selecting the optimal format for \gls{spmv} both on CPU and GPU platforms. This approach requires a transformation/compression on the input matrix to a fixed size ($128\times128$) image-like representation to be fed to the network, however reports the highest accuracy from all ($93\%$ and $90\%$ on CPU and GPU platforms respectively) at the expense of higher prediction time compared to the \gls{ml} alternatives.

Zhao et al.\cite{overhead_conscious} takes the format selection approach a step further proposing an overhead-conscious selection mechanism for \gls{spmv}-based applications that also takes into account the overheads from format conversion. By building several regression models reports accuracy up to $88\%$ and average runtime speedup of applications in the range of $1.14\times$ - $1.43\times$.

Our proposed auto-tuner manages to combine speed of \gls{ml} methods such as the ones proposed by Li et al. and the accuracy reported by Zhao et al. on a range of storage formats and across most of the key HPC platforms. Note however that even-though the datasets reported in previous work are unbalanced, none of the authors reports the achieved balanced accuracy therefore is not possible to make a fair comparison between contributions based on the accuracy alone.

\section{Conclusions and Further Work}
Selecting the optimal sparse matrix storage format is important for allowing applications to remain optimal across the available hardware architectures. However, the selection process is not a trivial task. \gls{ml} offers a systematic solution to this problem by approaching it as a classification task. In the case of the format selection, this problem can be categorized as a rare event prediction problem due to the imbalance observed in the data. By training, tuning and deploying an ensemble of \emph{decision trees}, we are able to accurately predict the optimum format to be used for the \gls{spmv} operation across the main HPC architectures. We find out that although most of the time the best option is to use \gls{csr}, in some cases the runtime performance is improved by orders of magnitude from switching to the optimal format. Finally, our proposed light-weight auto-tuning approach introduces overheads in the overall runtime of \gls{spmv} which are amortised quickly and within a few \gls{spmv} operations on average, with a more noticeable benefit to GPUs.

As a next step, we will explore ways of further improving the accuracy of our models either through balancing the dataset or other \gls{ml} methods such as \emph{gradient-boosted decision trees}. Furthermore, eliminating the requirement for manual feature extraction remains an avenue for further research.
\section*{Acknowledgment}
This research is part of the EPSRC project ASiMoV (EP/S005072/1). We used the ARCHER2 UK National Supercomputing Service (https://www.archer2.ac.uk) and the Cirrus UK National Tier-2 HPC Service at EPCC (http://www.cirrus.ac.uk), funded by the University of Edinburgh and EPSRC (EP/P020267/1), and Isambard~2 UK National Tier-2 HPC Service (http://gw4.ac.uk/isambard) operated by GW4 and the UK Met Office, and funded by EPSRC (EP/T022078/1).

\bibliographystyle{IEEEtran}  
\bibliography{bibliography/bibs.bib}

\end{document}